\documentclass[final,5p,times]{elsarticle}
\usepackage{graphicx}
\usepackage{subcaption}
\usepackage{epsfig}
\usepackage{siunitx}
\usepackage{booktabs}
\usepackage[T1]{fontenc}
\usepackage{multirow} 
\usepackage{tabularx}
\usepackage{array}
\usepackage[namelimits]{amsmath} 
\usepackage{amssymb} 
\usepackage{amsmath}            
\usepackage{amsfonts}            
\usepackage{mathrsfs} 
\usepackage{float}
\usepackage{cases}
\usepackage{wrapfig}
\usepackage{hyphenat}
\usepackage{multicol}
\usepackage{caption}
\usepackage[labelfont=bf, textfont=normalfont, justification=raggedright, singlelinecheck=false]{caption}
\usepackage{graphicx}
\usepackage{subcaption}
\usepackage{tabularx}  
\usepackage{amsmath}

\usepackage{amssymb}

\journal{arxiv}
\date{}

\begin{document}
\begin{frontmatter}



\title{A Feature Matching Method Based on Multi-Level Refinement Strategy \tnoteref{13}}


\tnotetext[13]{This work was supported in part by the National Natural Science Foundation of China (No. 62172190), National Key Research and Development Program(No. 2023YFC3805901), the "Double Creation" Plan of Jiangsu Province (Certificate: JSSCRC2021532) and the "Taihu Talent-Innovative Leading Talent" Plan of Wuxi City(Certificate Date: 202110). }
\author[label1]{Shaojie Zhang}
\ead{7213107006@stu.jiangnan.edu.cn}
\author[label1,label2]{Yinghui Wang\corref{cor1}}
\ead{wangyh@jiangnan.edu.cn}
\author[label1]{Jiaxing Ma}
\ead{2458098051@qq.com}
\author[label1]{Wei Li}
\ead{cs_weili@jiangnan.edu.cn}
\author[label1]{Jinlong Yang}
\ead{yjlgedeng@163.com}
\author[label1]{Tao Yan}
\ead{yantao.ustc@gmail.com}
\author[label1]{Yukai Wang}
\ead{ericwangyk22@163.com}
\author[label3]{Liangyi Huang}
\ead{lhuan139@asu.edu}
\author[label4]{Mingfeng Wang}
\ead{mingfeng.wang@brunel.ac.uk}
\author[label5]{Ibragim R. Atadjanov}
\ead{ibragim.atadjanov@gmail.com}
\cortext[cor1]{Corresponding author}
\affiliation[label1]{organization={ School of Artificial Intelligence and Computer Science, Jiangnan University},
            addressline={1800 Li Lake Avenue},
            city={wuxi},
            postcode={214122},
            state={Jiangsu},
            country={PR China}}
\affiliation[label2]{organization={ Engineering Research Center of Intelligent Technology for Healthcare, Ministry of Education},
            addressline={1800 Li Lake Avenue},
            city={wuxi},
            postcode={214122},
            state={Jiangsu},
            country={PR China}}
 \affiliation[label3]{organization={School of Computing and Augmented Intelligence, Arizona State University},
            addressline={1151 S Forest Ave},
            city={Tempe},
            postcode={8528},
            state={AZ},
            country={U.S}}
\affiliation[label4]{organization={Department of Mechanical and Aerospace Engineering, Brunel University},
            addressline={Kingston Lane},
            city={London},
            postcode={UB8 3PH},
            state={Middlesex},
            country={U.K}}    
\affiliation[label5]{organization={Tashkent University of Information Technologies named after al-Khwarizmi},
			addressline={ 108 Amir Temur Avenue},
			city={Tashkent},
		    postcode={100084},
			state={},
			country={Uzbekistan}}

\begin{abstract}
Feature matching is a fundamental and crucial process in visual SLAM, and precision has always been a challenging issue in feature matching. In this paper, based on a multi-level fine matching strategy, we propose a new feature matching method called KTGP-ORB. This method utilizes the similarity of local appearance in the Hamming space generated by feature descriptors to establish initial correspondences. It combines the constraint of local image motion smoothness, uses the GMS algorithm to enhance the accuracy of initial matches, and finally employs the PROSAC algorithm to optimize matches, achieving precise matching based on global grayscale information in Euclidean space. Experimental results demonstrate that the KTGP-ORB method reduces the error by an average of 29.92\% compared to the ORB algorithm in complex scenes with illumination variations and blur.\end{abstract}



\end{frontmatter}


\section{Introduction}
Feature matching is a fundamental and critical process in feature-based visual SLAM (Simultaneous Localization and Mapping) \cite{ref1}-\cite{ref3}. It directly impacts the precision, efficiency, and robustness of SLAM systems. Feature matching involves extracting significant structural features with physical meaning, such as feature points and lines, from two images. These similar or identical structures are then identified and aligned pixel-wise through feature matching, a process built on feature detection and descriptor construction. In visual SLAM, various feature detection methods ultimately converge on feature point detection. Commonly used classical feature point detection methods include SIFT \cite{ref4}, SURF \cite{ref5}, and ORB \cite{ref6}. Among them, SIFT and SURF require defining high-dimensional feature descriptors for extracting image features, resulting in high computational complexity. In contrast, ORB is generally faster than SIFT and SURF, making it more suitable for real-time implementation in SLAM systems. The ORB algorithm employs the rBRIEF descriptor \cite{ref20}, a 256-bit binary vector. The similarity between two feature points can be evaluated by calculating the Hamming distance between their descriptors. In the overall process, most visual SLAM systems typically adopt a two-stage matching scheme: initial coarse matching using brute force (BF) \cite{ref21} followed by RANSAC (Random Sample Consensus) algorithm to iteratively search for the optimal match from a set of feature matches that may contain redundant or even outlier pairs. Although these methods demonstrate good performance and robustness in feature-based visual SLAM, a single-level feature matching method often falls short in handling complex scenes. During the matching process, where thousands of feature points can be extracted from two images, leading to numerous potential match relationships, relying solely on local descriptors for feature matching inevitably results in ambiguity and a large number of false matches. To address this, researchers have proposed leveraging various pieces of information, such as local similarity in the Hamming space, local image structure in Euclidean space, and global grayscale information, to establish a multi-level matching strategy, aiming to eliminate ambiguity and false matches caused by insufficient information \cite{ref15}-\cite{ref17}. These methods primarily establish initial match relationships based on the similarity of local descriptors of feature points. Subsequently, they remove incorrect matches based on geometric constraints, with the elimination of incorrect matches mainly relying on resampling-based methods. However, methods based on resampling heavily depend on the accuracy of sampling. When there are many incorrect matches in the initial matching, the effectiveness of these methods is compromised. To address this issue, this paper integrates smoothness constraints into the matching process, precluding a large number of incorrect matches. Inspired by multi-level matching strategies \cite{ref15}-\cite{ref18}, we propose a feature matching method based on multi-level refinement to enhance feature matching in complex scenes.

The innovative points of this article can be summarized as follows:

(1) A novel multi-level refinement matching strategy is designed. Inspired by the abundance of matching points around feature points due to motion smoothness, the strategy incorporates the constraints of local image motion smoothness into the multi-level refinement strategy. Combining feature matching with fast data association in Hamming space addresses the low accuracy issue in initial matches based on resampling methods.

(2) A new KTGP-ORB model is proposed. This model supports feature extraction and matching, delivering satisfactory results in terms of accuracy even in complex environments.

\begin{figure*}[t]
    \centering
    \includegraphics[height=4.5in]{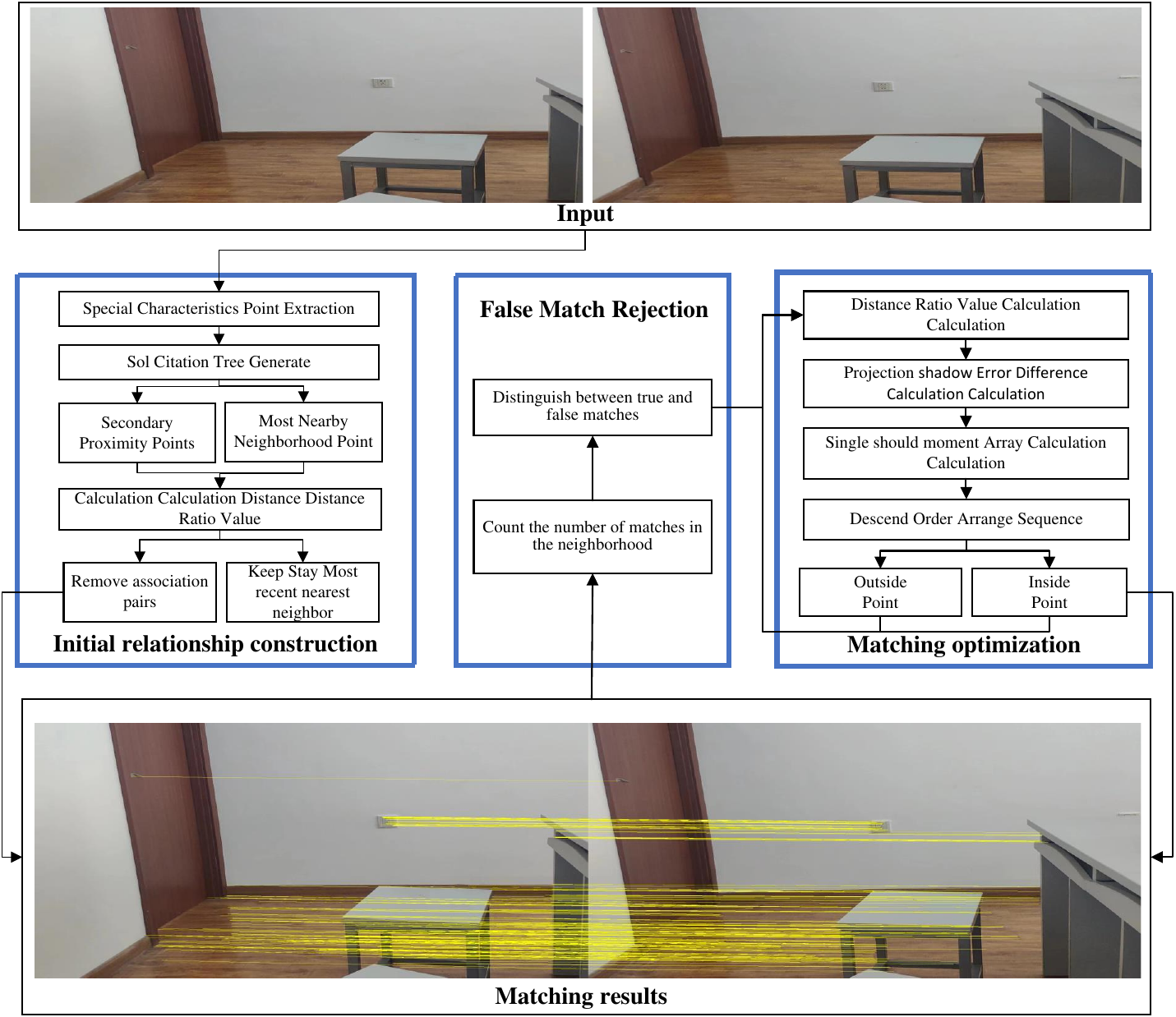}
\caption{Methodological Framework.}
    \label{Fig1}
\end{figure*}

\section{RELATED WORK}
During the matching process, a significant number of erroneous and false matches are inevitable. In 2017, Bian et al. proposed the GMS algorithm \cite{ref7}, which employs statistical methods to count the number of matches within local grid regions. The algorithm determines whether all matches within a grid region are correct based on the quantity of matches. In 2018, Chen et al. \cite{ref8} optimized GMS by reducing the grid size from 9 to 5 and the rotation matrix calculations from 7 to 3, albeit at the expense of rotational invariance. Fischler et al. introduced the Random Sample Consensus (RANSAC) algorithm in 1982 \cite{ref9}, which has long been recognized as the most universally applicable and effective method for error match filtering. RANSAC selects samples randomly from input data to compute the best model, with samples satisfying this model termed inliers or correct matches. Various improved forms of RANSAC, such as MLESAC \cite{ref10}, PROSAC \cite{ref11}, SCRAMSAC \cite{ref12}, USAC \cite{ref13}, have been proposed, collectively known as resampling-based methods. These methods heavily depend on the accuracy of sampling, and when there are too many erroneous matches in the initial matching, the required number of samples significantly increases, resulting in reduced efficiency.

In 2019, Zhu et al. \cite{ref14} proposed GMS-RANSAC based on improved grid motion statistical features, utilizing distance similarity to eliminate outliers and enhance accuracy, albeit at the cost of increased runtime. When dealing with complex scenes, single-level feature matching methods often fail to meet practical requirements. Designing a fast and accurate strategy for handling mismatch issues remains a challenge, especially in visual SLAM systems. Ye et al. \cite{ref15} proposed the MP-ORB matching method to enhance matching accuracy by combining K Nearest Neighbors (KNN), neighbor ratio, bidirectional matching, cosine similarity (CS), and PROSAC. In addressing irregular dynamic changes in brightness and contrast caused by non-uniform illumination and the random appearance of textureless areas, Sun et al. \cite{ref16} introduced a multi-stage matching module consisting of KNN, threshold filtering, feature vector norms, and RANSAC to eliminate mismatches. Sun et al. \cite{ref17} proposed a KTBER multi-stage matching technique based on the similarity of feature descriptors in Hamming space and Euclidean space, as well as the global grayscale information of feature pairs. Later, Sun et al. \cite{ref18} introduced the RTC (Ratio-test Criterion) technique \cite{ref19} to compare the ratio of nearest neighbor distance to the second nearest neighbor with a threshold, presenting a KRCR multi-level matching strategy composed of KNN, threshold filtering, RTC, cosine similarity, and RANSAC.

Most of the aforementioned methods consider the quality of matches influenced by the invariance and distinctiveness of features in Hamming and Euclidean spaces. Initial matches are established based on this, with error match removal relying primarily on resampling methods like RANSAC or PROSAC. However, matches established in this way are susceptible to noise, blur, and occlusion, leading to numerous erroneous match\\es. Resampling-based methods heavily rely on the accuracy of sampling, and when there are many erroneous matches in the initial matching, the effectiveness of such methods is significantly compromised. Motion smoothness constraint posits that correspondence sets caused by motion smoothness cannot occur randomly. Therefore, by simply calculating the number of matches near feature points, true and false matches can be distinguished. This article integrates smoothness constraints into the matching process, precluding a large number of erroneous matches, and further enhances the ability to effectively eliminate erroneous and low-quality matches in complex scenes through a multi-level refinement strategy for feature matching.

\section{METHODOLOGY}
In complex scenes, a multitude of erroneous matches in the initial matching phase can lead to a decrease in the accuracy of resampling-based methods, thereby impacting the overall precision of the multi-level matching process. Therefore, to enhance the accuracy of matching in complex workspaces, the proposed feature matching method based on a multi-level refinement strategy follows the technical roadmap depicted in Figure 1. This roadmap encompasses three major aspects: initial correspondence generation, false match removal, and matching optimization.

\subsection{Initial Correspondence Generation}
Utilizing the KNN algorithm, a large set of one-to-two associated feature pairs is rapidly generated. Subsequently, a threshold filtering method is employed to evaluate the matching quality based on the ratio of the nearest neighbor distance to the second nearest neighbor distance, thereby enhancing the efficiency of establishing initial data associations between two given feature sets. For feature matches that meet the specified conditions, only the nearest neighbor points are retained, transforming the one-to-two associated feature pairs into one-to-one matching pairs, providing an initial set of matching pairs for subsequent removal of erroneous matches.

KNN, in the matching process, selects K points most similar to a feature point. If the differences among these K points are significant, the most similar point is chosen as the matching point. In this study, K is set to 2 for the nearest neighbor matching. By selecting two nearest neighbors for each feature point in the target image from the reference image, a rapid construction of initial data associations between the two given feature sets is achieved, generating a large set of one-to-two associated feature pairs.

Let $P_{ti}=\begin{Bmatrix} p_{t1},p_{t2},p_{t3},\ldots, p_{ta}\end{Bmatrix}$ represent the sampling set of feature points in the target image $I_{t+1}$, and $P_{rj}=\begin{Bmatrix}
	p_{r1},\ldots,p_{rb}
	\end{Bmatrix} $ represent the template set of feature points in the reference image $I_{t}$. A Kd-tree algorithm is employed to generate an index tree for feature descriptors, enabling a fast search for $K$ nearest neighbors from $p_{t1}$ to $P_{rj}$. Since this study uses $K=2$, for each feature in $P_{rj}$ the nearest neighbor point and the second nearest neighbor point with the smallest Hamming distances are found in $P_{rj}$, as expressed in Equation (1).

\begin{equation}
	\label{eq1}
	d(p_{t_{i}},p_{r_{j}})=\sum (p_{t_{i}}(F)\oplus p_{r_{j}}(F))\\
1\leq i\leq a, 1\leq j\leq b     
\end{equation}

In the above, a,b represent the respective numbers of feature points in images $I_{t+1}$ and $I_{t}$, $\oplus$ denotes the XOR operation, and  $P_{ti}$ and $P_{rj}$ represent 256-bit binary vectors.

Therefore, for each feature point  $P_{ti}$ in the set  $P_{ti}$, the nearest neighbor point and the second nearest neighbor point can be found in the set $P_{rj}$, as shown in Equation (2). 
\begin{multline}
	\label{eq2}
	P_{i,r\rightarrow }=\\\left \{ (p_{1,s1},p_{1,s2}),\ldots , (p_{a,s1},p_{a,s2})\right \}
\left \{ p_{1,s1},p_{2,s1}, \ldots ,p_{a,s1}\right \}  
\end{multline}

In this context, $ \left \{  p_{1,s1},p_{2,s1}, \ldots ,p_{a,s1} \right \}$ represents the set of nearest neighbors corresponding to  $P_{ti}$ in the set $P_{rj}$, and $ \left \{  p_{1,s2}, \\\ldots  \right \}$ represents the set of second nearest neighbors corresponding to  $P_{ti}$ in the set $P_{rj}$.

The nearest neighbor point and the second nearest neighbor point form a one-to-two associated feature pair obtained through KNN matching. Since there may be many non-matching pairs between the two feature sets, a threshold filtering method is employed for rapid removal of these non-matching pairs, transforming the one-to-two associated feature pairs into one-to-one matching pairs.

Assuming the distances from point  $P_{ti}$ to points $p_{1,s1}$ and $p_{1,s2}$ are $d_1 (P_{t1}, p_{1,s1} ) $ and $d_2 (P_{t1}, p_{1,s2} ) $ respectively, the ratio of the nearest neighbor distance to the second nearest neighbor distance, i.e., the ratio of $d_1$ to $d_2$, is computed as shown in Equation (3).
\begin{equation}
	\label{eq3}
	W= d_{1}(p_{t_{1}},p_{1,s1})/d_{2}(p_{t_{1}},p_{1,s2})
\end{equation}
When $W$ is less than the given threshold $T_w$, the one-to-two associated pairs $(P_{t1}, p_{1,s1} )$ and $(P_{t1}, p_{1,s2} )$ are retained, transforming into one-to-one associated pairs by keeping $(P_{t1}, p_{1,s1} )$. Conversely, if this ratio exceeds the threshold $T_w$, both associated pairs between $P_{t1}$ and $(P_{1,s1}, p_{1,s2} )$ are entirely removed.

\subsection{False Match Removal}
After obtaining the initial correspondence relations, the characteristics of motion smoothness causing a higher density of matching points around matched feature points are utilized to eliminate some non-matching pairs. This is mainly done to retain a sufficient number of matching pairs in challenging scenes. In the threshold filtering process, a larger threshold is often preferred, leading to the possibility of a considerable number of non-matching pairs remaining in the remaining matching pairs after threshold filtering, significantly reducing the effectiveness of resampling-based methods.

Inspired by the GMS algorithm, which posits that the lack of clearly correct matches is not due to a low quantity of matching pairs but rather the difficulty in distinguishing between correct and incorrect matches, motion smoothness constraint is transformed into statistical matching pairs' neighborhood matching quantity. This is used as a basis for determining true and false matches. Since the feature points in the neighborhood of correct matches often maintain the consistency of motion smoothness, evaluating the number of matching pairs in the feature point's neighborhood being assessed can differentiate between correct and incorrect matches. If the number of matches in the neighborhood is less than a given threshold, the matching pair is considered an incorrect or low-quality match; otherwise, it is deemed a correct match. The decision process is illustrated in Figure 2, where if there are other matching pairs in the neighborhood of a pair of matching points, the likelihood of this pair being correct is relatively high.

\begin{figure*}[t]
    \centering
    \includegraphics[height=2.5in]{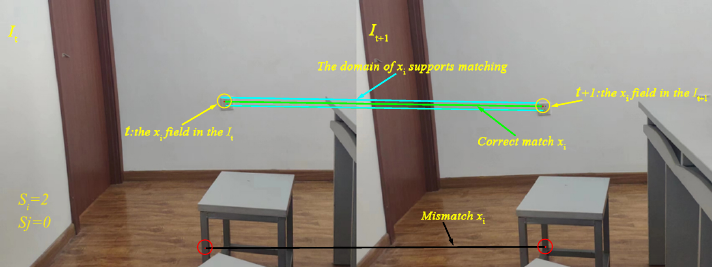}
\caption{Examples of removing false matches.}
    \label{Fig2}
\end{figure*}

Assuming two input images are denoted as $(I_{t}, I_{t+1} )$, each having ${N,M}$ feature points, let $X={x_1,x_2, ,x_i,\cdots,x_N }$ represent the set of all matching pairs from the image $I_t$ to $I_{t+1}$. The region ${t,t+1}$ pertains to the neighborhood of matching pairs $x_i$ in the images ${I_t,I_(t+1) }$, where each neighborhood has ${n,m}$ supporting matching pairs (excluding the original matching pair). The quantity of matching pairs within each neighborhood is also referred to as the neighborhood support, and its calculation is expressed in Equation (4).
\begin{equation}
	\label{eq4}
	S_{i}= \left | X_{i} \right |-1
\end{equation}
Here, $X_i\epsilon X$ represents the subset of matches between the neighborhood ${t,t+1}$ corresponding to the matching pair $x_i$, and $S_i$ is the neighborhood support for the matching pair $x_i$. The formula indicates the support excluding the matching pair $x_i$ itself. Consequently, the calculation for the scenario depicted in Figure 2 would yield $S_i=2,S_j=0$.

\subsection{Matching Optimization}
After removing a significant number of false matches, matching pairs that remain undergo matching optimization to further enhance the accuracy of feature matching. The quality of matching pairs corresponds to the optimal homography transformation between the feature points. Therefore, this paper adopts the Progressive Sample Consensus (PROSAC) method [11] to estimate the optimal homography matrix, distinguish inliers from outliers, and further eliminate low-quality and redundant matches. PROSAC is an optimization of the classical RANSAC method, where the PROSAC algorithm samples from an increasingly optimal set of corresponding points. Compared to the classical RANSAC algorithm, which uniformly samples from the entire set, PROSAC can save computation and improve runtime speed. The core of the PROSAC algorithm involves pre-sorting the sample points, selecting suitable sample point pairs for estimating the matching model, reducing the randomness of the algorithm, and achieving high model accuracy, thereby reducing the number of algorithm iterations.

According to the assumptions of the PROSAC algorithm, the higher the similarity between data points, the higher the probability of being an inlier. To this end, an evaluation function q(u) is introduced to represent the probability of a data point becoming an inlier. Considering the entire set $U_N$ of N data points, the calculation for sorting within $U_N$ is expressed in Equation (5).
\begin{equation}
	\label{eq5}
	i< j\Rightarrow q(u_{i})\geq q(u_{j})\quad u_{i},u_{j}\epsilon U_{N}
\end{equation}

For each pair of feature points in the image, the quality of the feature point matching is measured using the ratio $\beta$ of Euclidean distances, as calculated in Equation (6).
\begin{equation}
	\label{eq6}
	\beta =\frac{d_{min}}{d_{min2}}
\end{equation}
Here, $d_min$ is the minimum Euclidean distance, $d_min2$ is the second minimum Euclidean distance, and a smaller $\beta$  indicates a higher probability $q(u)$ of being an inlier, signifying better matching quality.

Subsequently, the initial matching pairs are sorted in descending order based on the evaluation function. Let $p{u_i }$ denote the probability of $u_i$ being a correct match within the sorted subset, and a correlation assumption between this probability and the evaluation function is expressed in Equation (7).	
\begin{multline}
	\label{eq7}
	i< j\Rightarrow q(u_{i})\geq q(u_{j}) \Rightarrow p(u_{i})\geq p(u_{j})\quad \\u_{i} ,u_{j}\epsilon U_{N}   
	\left \{ M_{j}\right \}\begin{matrix}
	T_{N}\\ i=1
	\end{matrix}
\end{multline}

The top $n$ data points with the maximum correlation function values are then selected as the set $U_n$, and m data points are sampled from this set, forming the set $M$. Based on the quality of the samples, the sequence of $T_N$ samples taken from the data set $U_N$ is denoted as $\left \{ M_{i} \right \}_{i=1}^{T_{N}}$. If the sequence $\left \{ M_{i} \right \}_{j=1}^{T_{N}}$ is sorted according to the evaluation function, it holds that:
\begin{equation}
	\label{eq8}
	i< j\Rightarrow q(u_{i})\geq q(u_{j}) \Rightarrow p(u_{i})\geq p(u_{j})\Rightarrow q(M_{i})\geq q(M_{j})
\end{equation}

After sorting the matching quality of feature points, every set of four feature points is grouped, and the sum of qualities for each group is calculated and sorted. The top four groups of matching points are selected to calculate their homography matrix. These four groups of points are then removed, and the remaining points are used to calculate the corresponding projected points based on the homography matrix. The projection error between these points is computed and compared to a predefined inlier threshold. If it is less than the threshold, the points are considered inliers; otherwise, they are outliers. The number of obtained inliers is compared to a predefined inlier count threshold. If it is greater than the threshold, the inlier count is updated to the current value; otherwise, the iteration continues until the optimal match is obtained. In this process, the default inlier threshold and inlier count threshold from OpenCV are used for discrimination.

\section{Experimental Setup and Result Analysis}
This section validates the feasibility of the proposed method through experiments, including the selection of evaluation criteria, dataset choices, experimental results, parameter variation experiments, and comparative analyses with existing methods. The experiments were conducted on a laptop with an Intel i7-4710MQ processor, 8GB RAM, 500GB hard disk capacity, and Ubuntu 18.04 operating system.
\subsection{Dataset Selection}
We utilized two real-world scene images collected by ourselves to demonstrate the experimental results. Additionally, we chose the Oxford Visual Geometry Group's Optical Images Dataset \cite{ref22} for comparative experiments. This dataset provides numerous image pairs with various changes in lighting, blur, and other factors. Specifically, we selected the Leuven dataset (lighting changes) and the Bikes dataset (blur) as representative examples of complex scenes. In each dataset, the first image is matched with the remaining five images, generating five image pairs with gradually decreasing quality. These datasets pose significant challenges for feature extraction and matching, serving as effective tests for the proposed method's validity. 
\subsection{Evaluation Criteria Selection}
This paper primarily employs evaluation metrics such as the Number of Matchings (NM), Repeatability (REP) [23], Mean Error (ME), and Root Mean Square Error (RMSE) to compare the accuracy of the proposed method.
	
Repeatability (REP) measures the number of times different instances of the same object or feature point are matched. Higher repeatability indicates better algorithm stability. The calculation of repeatability is shown in Equation (9).
\begin{equation}
	\label{eq9}
	REP= \frac{n_{matched}}{n_{detected}}
\end{equation}
Here, $n_matched$ is the number of matched feature points, and $n_detected$ is the total number of detected feature points across all images.

Mean Error (ME) represents the average distance error between all correctly matched feature point pairs. A smaller mean error indicates higher algorithm accuracy. The calculation of mean error is expressed in Equation (10).
\begin{equation}
	\label{eq10}
	ME= \frac{1}{n_{matched}}\sum_{i=1}^{n_{matched}}\left \| p_{i}-q_{i} \right \|
\end{equation}
Where $p_i$ and $q_i$ are the coordinates of the ith matched feature point in the two images.

Root Mean Square Error (RMSE) signifies the root mean square of the distance errors between all correctly matched feature point pairs. A smaller RMSE indicates higher algorithm accuracy. The calculation of RMSE is given in Equation (11).
\begin{equation}
	\label{eq11}
	RMSE= \sqrt{\frac{1}{n_{matched}}\sum_{i=1}^{n_{matched}}\left \| p_{i}-q_{i} \right \|^{2}}
\end{equation}
\subsection{Experimental Results}
As shown in Figure 3(a), two images collected by ourselves were used as input for the algorithm. Feature points were extracted from the input images, and the extraction results are depicted in Figure 3(b), with the extracted feature points represented by red dots.

\begin{figure}[t]
    \centering
	\begin{subfigure}{0.7\linewidth}
		\centering
		\includegraphics[width=0.95\linewidth]{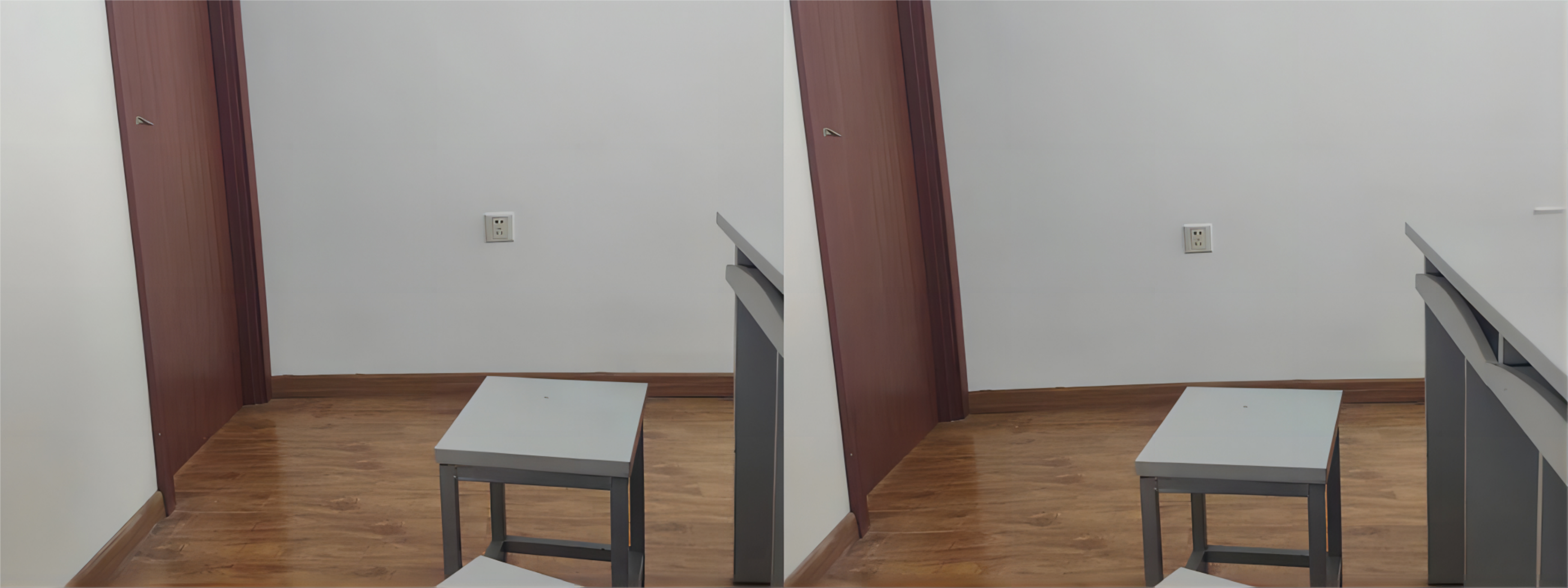}
		\caption{Original image}
		\label{Fig31}%
	\end{subfigure}
	\centering
	\begin{subfigure}{0.7\linewidth}
		\centering
		\includegraphics[width=0.95\linewidth]{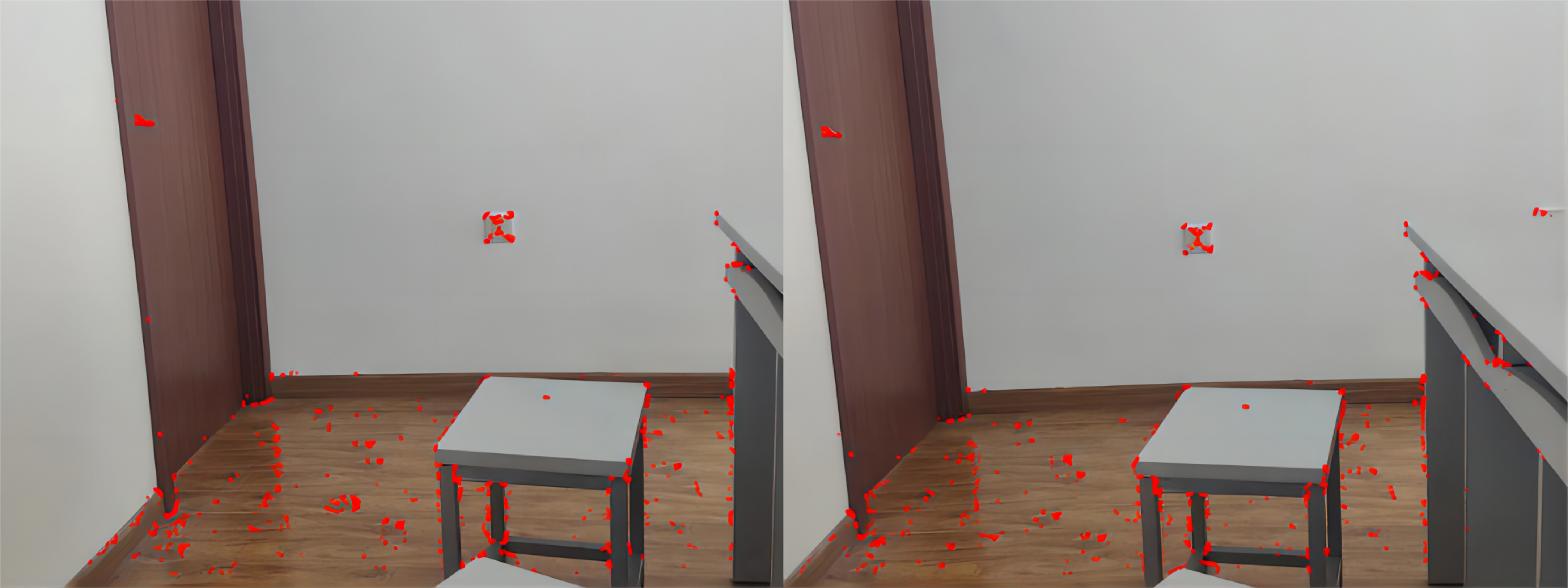}
		\caption{Feature point extraction results}
		\label{Fig32}
	\end{subfigure}
	\caption{Two self-collected input images and their feature point extraction results.}
    \label{Fig3}
\end{figure}

After feature extraction, a multi-level refinement matching was performed on the extracted feature points. All matching pairs are marked with yellow lines in each image pair, as shown in Figure 4. The KTGP-ORB model initially establishes a one-to-two data association between the feature points in the two images rapidly using the KNN algorithm, as illustrated in Figure 4(a). To convert the one-to-two data association into a one-to-one association, the TF technique is employed to find the optimal match within the one-to-two data association, forming a one-to-one matching relationship, as shown in Figure 4(b). Subsequently, in Figure 4(c), to eliminate low-quality and erroneous matches after TF, the GMS algorithm is applied to further filter out incorrect matches. The GMS algorithm considers the matching quantity within the neighborhood of matching points, transforming a higher quantity of feature point matches into higher-quality matches, effectively distinguishing correct matches from incorrect ones. Finally, to obtain the optimal matches, PROSAC is employed for the last filtering in the KTGP-ORB multi-level refinement matching model. The final results are shown in Figure 4(d). The experimental results intuitively validate the effectiveness of the KNN-TF-GMS-PROSAC (KTG\\P) multi-level refinement matching model.

\begin{figure}[t]
    \centering
	\begin{subfigure}{0.7\linewidth}
		\centering
		\includegraphics[width=0.95\linewidth]{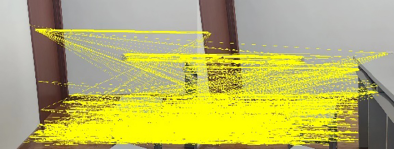}
		\caption{KNN}
		\label{Fig41}
	\end{subfigure}
	\centering
	\begin{subfigure}{0.7\linewidth}
		\centering
		\includegraphics[width=0.95\linewidth]{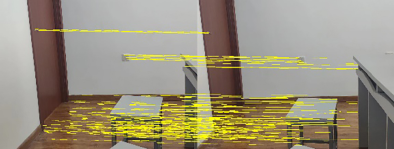}
		\caption{KNN+TF}
		\label{Fig42}
	\end{subfigure}
	\centering
	\begin{subfigure}{0.7\linewidth}
		\centering
		\includegraphics[width=0.95\linewidth]{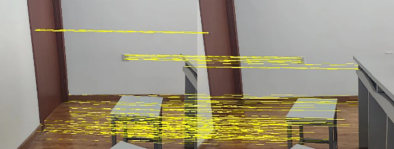}
		\caption{KNN+TF+GMS }
		\label{Fig43}
	\end{subfigure}
	\centering
	\begin{subfigure}{0.7\linewidth}
		\centering
		\includegraphics[width=0.95\linewidth]{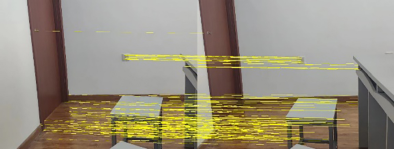}
		\caption{KNN+TF+GMS+PROSAC}
		\label{Fig44}
	\end{subfigure}
	\caption{Example of matching results.}
    \label{Fig4}
\end{figure}
To further validate the effectiveness of the KNN-TF-GMS-PROSAC (KTGP) multi-level refinement matching model in complex scenes with lighting changes, blurriness, and other challenging conditions, comprehensive performance evaluations are conducted on the Leuven dataset (lighting changes), Bikes dataset (blurriness), and Boat dataset (camera rotation). As shown in Figure 5, the experimental results of multi-level refinement matching on five pairs of images from the Leuven dataset and Bikes dataset are presented, with all matching pairs marked with yellow lines in each image pair. Combining the data in Table 1, it is observed that the KTGP multi-level refinement matching technique can effectively eliminate false matches and redundant matches in complex scenes with lighting changes and blurriness. As the severity of lighting changes and blurriness increases, the number of extracted feature points decreases, leading to an increase in false matches. However, the KTGP multi-level refinement matching model can still selectively filter out correct matches, and the quantity of correct matches decreases with the increasing severity of the variations. The matching results intuitively demonstrate the effectiveness of the multi-stage KNN-TF-GMS-PROSAC (KT\\GP) multi-level refinement matching model in the feature matching stage.

\begin{figure}[H]
    \centering
	\begin{subfigure}{0.45\linewidth}
		\centering
		\includegraphics[width=0.95\linewidth]{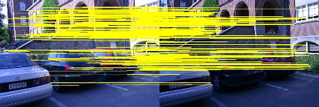}
		\caption{Leuven 1-2}
		\label{Fig51}
	\end{subfigure}
	\centering
	\begin{subfigure}{0.45\linewidth}
		\centering
		\includegraphics[width=0.95\linewidth]{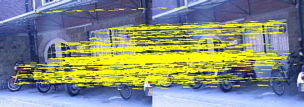}
		\caption{Leuven 1-2}
		\label{Fig52}
	\end{subfigure}
	\centering
	\begin{subfigure}{0.45\linewidth}
		\centering
		\includegraphics[width=0.95\linewidth]{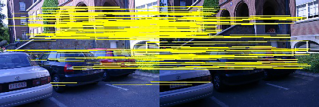}
		\caption{Leuven 1-3 }
		\label{Fig53}
	\end{subfigure}
	\centering
	\begin{subfigure}{0.45\linewidth}
		\centering
		\includegraphics[width=0.95\linewidth]{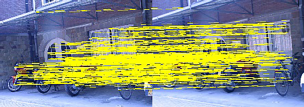}
		\caption{Leuven 1-3}
		\label{Fig54}
	\end{subfigure}
	\centering
	\begin{subfigure}{0.45\linewidth}
		\centering
		\includegraphics[width=0.95\linewidth]{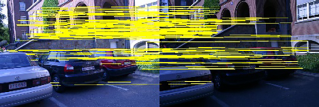}
		\caption{Leuven 1-4}
		\label{Fig55}
	\end{subfigure}
	\centering
	\begin{subfigure}{0.45\linewidth}
		\centering
		\includegraphics[width=0.95\linewidth]{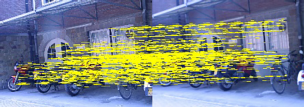}
		\caption{Leuven 1-2}
		\label{Fig56}
	\end{subfigure}
	\centering
	\begin{subfigure}{0.45\linewidth}
		\centering
		\includegraphics[width=0.95\linewidth]{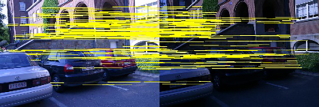}
		\caption{Leuven 1-2}
		\label{Fig57}
	\end{subfigure}
	\centering
	\begin{subfigure}{0.45\linewidth}
		\centering
		\includegraphics[width=0.95\linewidth]{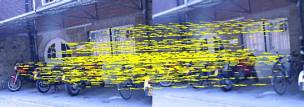}
		\caption{Leuven 1-3 }
		\label{Fig58}
	\end{subfigure}
	\centering
	\begin{subfigure}{0.45\linewidth}
		\centering
		\includegraphics[width=0.95\linewidth]{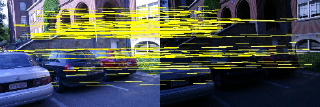}
		\caption{Leuven 1-3}
		\label{Fig59}
	\end{subfigure}
	\centering
	\begin{subfigure}{0.45\linewidth}
		\centering
		\includegraphics[width=0.95\linewidth]{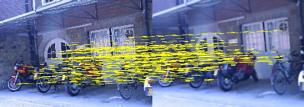}
		\caption{Leuven 1-4}
		\label{Fig510}
	\end{subfigure}
	\caption{Example of matching results.}
    \label{Fig5}
\end{figure}

\begin{table}[!hptb]
	\centering
	\caption{The method described in this text on the matching number over five pairs of images from the Leuven dataset and the Bikes dataset.}
	\setlength{\tabcolsep}{15pt}
	\begin{tabular} {lcc}
	  \toprule
	  
     Image Pairs  & Leuven & Bikes  \\
	  \midrule

	1-2	& 785  &928\\
	1-3	&546	&767\\
	1-4	&405	&456\\
	1-5	&352	&330\\
	1-6	&309	&198\\
	
	  \bottomrule
	\end{tabular}
	
	\label{tab1}
  \end{table}    

  \subsubsection{Parameter Variation Experiment}
  This section initially demonstrates the impact of the threshold $T_w$ during the Threshold Filtering (TF) process on the number of matches retained after threshold filtering. This helps us find the optimal threshold that yields the best results. The method employs the controlled variable approach \cite{ref15} to determine the threshold. In challenging scenarios with lighting changes and blurriness, the threshold $T_w$ is adjusted within the range of 0.1 to 0.9. The experiment records the percentage $Q$ of retained matches relative to the total number of original matches. The results are presented in Figures 6 and 7.
  
  For lighting changes and blurriness, five pairs of images from the Leuven dataset and Bikes dataset are chosen as the test image pairs. The decision for the threshold $T_w$ is made by analyzing how Q varies with $T_w$. From Figures 6 and 7, it is evident that the general trend in both scenarios is that the retention percentage $Q$ increases with an increase in the threshold $T_w$, and higher scene complexity leads to a lower retention percentage $Q$. However, using a larger threshold for threshold filtering retains more matching pairs but also increases the number of low-quality and erroneous matches. Simultaneously, using a smaller threshold enhances matching accuracy but may filter out some potentially correct matches.
  
  As shown in Figures 6 and 7, when $T_w \leq  0.3$, the retention percentage $Q$ is zero or close to zero for image pairs of different complexity levels in both scenarios. For example, at a threshold of 0.1, image pairs 1-4, 1-5, and 1-6 in both scenarios exhibit near-zero retention percentage $Q$. When $0.3 < T_w \leq  0.5$, the retention percentage $Q$ remains low for image pairs with different degrees of blurriness. For instance, at a threshold of 0.4, image pairs 1-5 and 1-6 in the Bikes dataset still show low retention percentages $Q$. However, when $T_w= 0.6$, all image pairs still retain an appropriate number of relatively correct matching pairs after threshold filtering. This indicates that $T_w$ must be greater than or equal to 0.6 to achieve precision meeting practical requirements.
  
  Since a larger threshold weakens the constraint, retaining more matching pairs but also more erroneous matches, a larger threshold is not necessarily better. Therefore, this study sets the threshold $T_w$ to 0.66.
  
  \begin{figure}[H]
    \centering
    \includegraphics[height=2.5in]{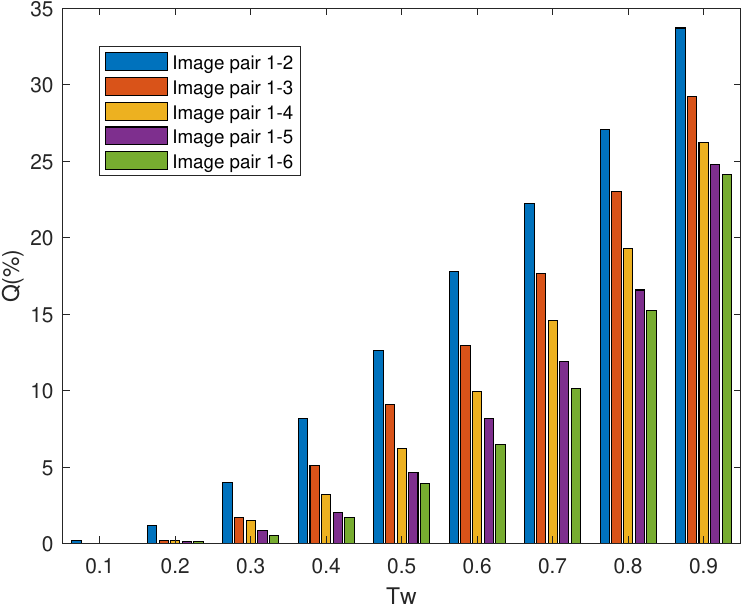}
\caption{Results of the retention percentage for five image pairs on the Leuven dataset under thresholds from 0.1 to 0.9.}
    \label{Fig6}
\end{figure}

\begin{figure}[H]
    \centering
    \includegraphics[height=2.5in]{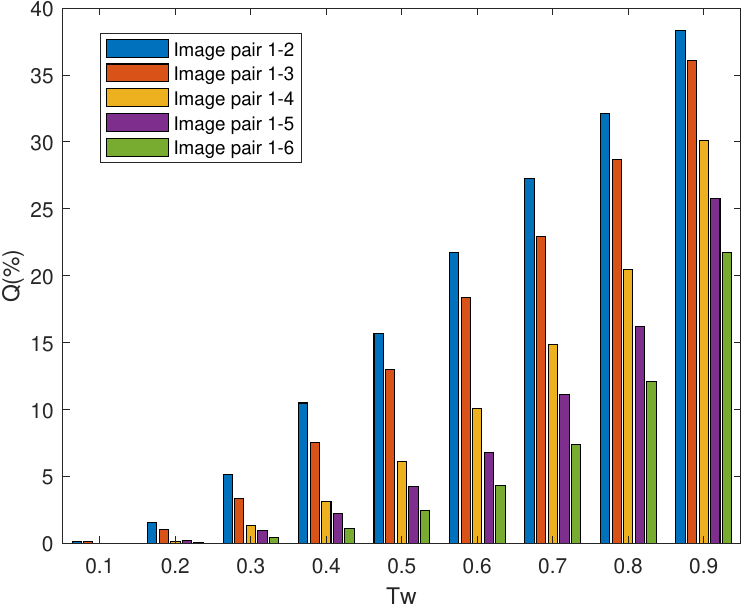}
\caption{Results of the retention percentage for five image pairs on the Bikes dataset under thresholds from 0.1 to 0.9.}
    \label{Fig7}
\end{figure}
  After determining the threshold for threshold filtering, we will continue to demonstrate the impact of the threshold on the number of matches after false match elimination. This will help choose an appropriate threshold. Similarly, the controlled variable method is employed. In challenging scenarios with lighting changes and blurriness, the threshold is adjusted in the range of 1 to 9. The experiment records the remaining number of matches NM (knn-tf-gms) after false match elimination and the difference between NM and the final number of matches. The results are presented in Figures 8 and 9.
  \begin{figure}[t]
    \centering
    \includegraphics[height=2.5in]{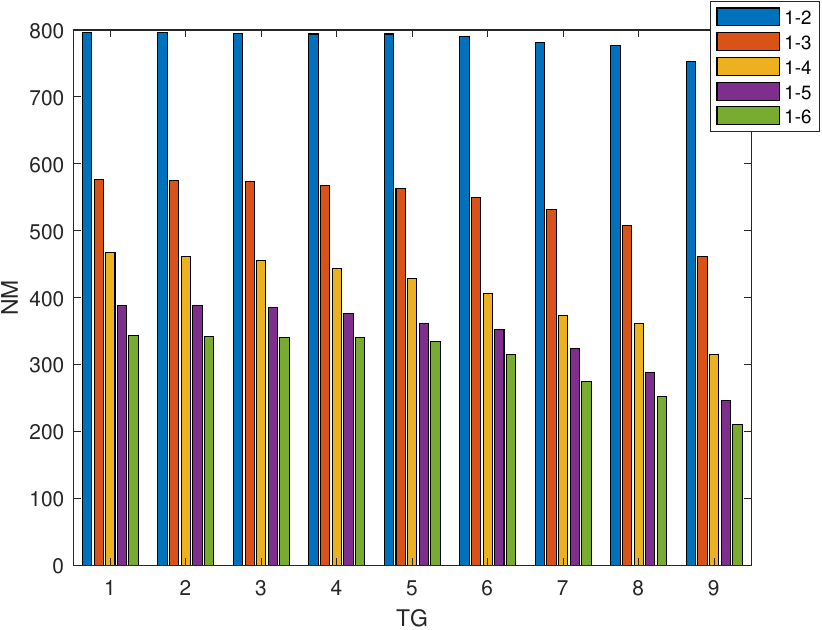}
\caption{NM results for five image pairs on the Leuven dataset under thresholds from 1 to 9.}
    \label{Fig8}
\end{figure}

\begin{figure}[t]
    \centering
    \includegraphics[height=2.5in]{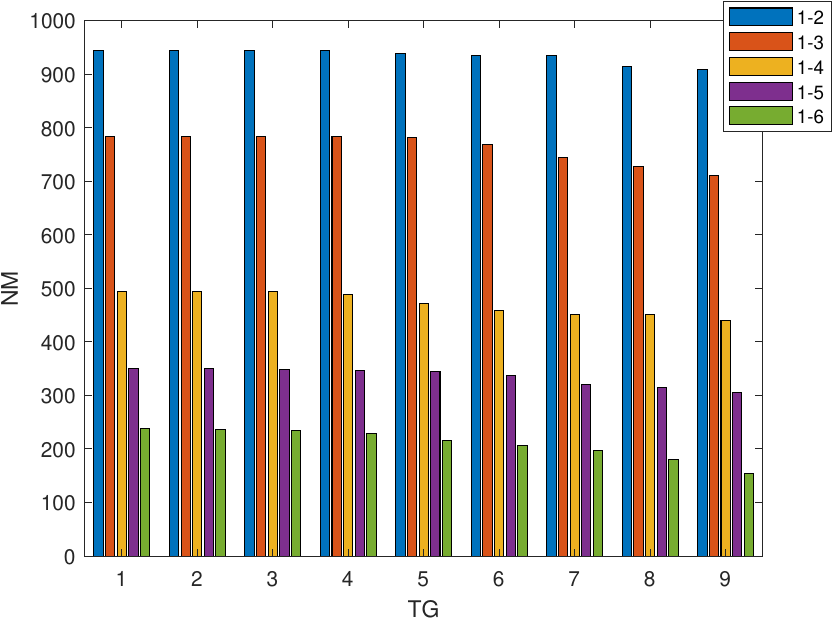}
\caption{NM results for five image pairs on the Bikes dataset under thresholds from 1 to 9.}
    \label{Fig9}
\end{figure}

  Similarly, five pairs of images from the Leuven dataset and Bikes dataset were selected as test image pairs. From Figures 8 and 9, it can be observed that the general trend in both scenarios is that the remaining number of matches, NM, decreases with the increase in the threshold $T_G$, and the complexity of the scene correlates with a lower NM. However, the reduction in NM with the increase in the threshold $T_G$ is relatively small. Therefore, relying solely on the general trend is insufficient for deciding on the threshold $T_G$. To address this, considering the difference between NM and the final number of matches, an analysis of the average difference in both scenarios is performed. This is to determine the appropriate threshold $T_G$, denoted as Avg, by analyzing the average change in the difference between NM and the final number of matches for the five image pairs. When Avg is large, it indicates that the false matches removed in the false match elimination stage are few, leaving more false matches in the initial matches for the subsequent matching optimization stage. Conversely, when Avg is small, although it indicates that the false matches removed in the false match elimination stage are more, the difference with the matching optimization after is small, but it may also mean that some potentially correct matches are excluded. The experimental results are shown in Figure 10. It can be seen that in both scenarios, the general trend is that Avg decreases with the increase in the threshold $T_G$, and Avg remains constant for thresholds 6 and 7. However, since using a smaller threshold to eliminate false matches retains more matches but also preserves more false and low-quality matches, it results in a larger Avg. On the other hand, using a larger threshold to eliminate false matches removes more false matches but may also filter out some correct matches. For example, when the threshold is between 6 and 7, Avg remains constant, but when > 7, Avg begins to decrease again. This suggests that after > 7, some correct matches are filtered out, leading to a subsequent decrease in the stable Avg. Combining this with the general trend of the remaining number of matches NM decreasing with the increase in the threshold $T_G$, the paper aims to retain a relatively large number of remaining matches NM. Therefore, the threshold $T_G$ is set to 6.
  \begin{figure}[t]
    \centering
    \includegraphics[height=2.5in]{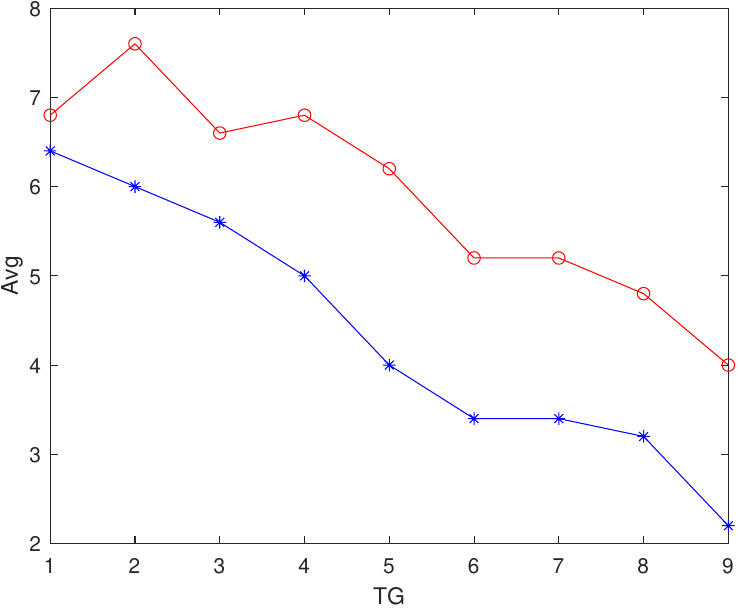}
\caption{Average difference between NM and the final matching number for five image pairs on the Leuven and Bikes datasets.}
    \label{Fig10}
\end{figure}
\subsubsection{Comparative Analysis of Methods}
In this section, an experimental analysis is conducted on models at different stages to evaluate the specific effects of the KNN-TF-GMS-PROSAC (KTGP) multi-stage refinement matching model. The models at different stages mainly include different matching techniques, such as KNN, TF, GMS, and PROSAC. To simplify the expression of different structured models in the experimental analysis, several abbreviations are provided based on the different matching techniques, as shown in Table 2.

\begin{table}[!hptb]
	\centering
	\caption{Abbreviations for models at different stages.}
	\setlength{\tabcolsep}{15pt}
	\begin{tabular} {lc} 
	  \toprule
	  
	  Abbreviation   & Matching Technique  \\
	  \midrule
	  KT-ORB	&KNN+TF\\
	  KTG-ORB	&KNN+TF+ GMS\\
	  KTGP -ORB	&KNN+TF+GMS+ PROSAC\\
	
	  \bottomrule
	\end{tabular}
	
	\label{tab2}
  \end{table}   
  In this section, the Leuven dataset and Bikes dataset, featuring challenging scenarios with lighting variations and blurriness, are utilized. The first image pair from each dataset is selected as sample images for feature extraction and matching to assess the performance of models at different stages. The experimental results are illustrated in Figure 8. The KT-ORB model effectively reduces some erroneous matches but may overlook others. By incorporating GMS after KNN and TF for feature matching, the KTG-ORB model further eliminates mismatches, enhancing the ratio of correct matches. In comparison to the aforementioned models, the KTGP-ORB model not only eliminates false matches but also removes redundant matches, resulting in an appropriate number of high-quality matches in challenging scenarios with lighting variations and blurriness.
  
  \begin{figure}[!hptb]	
	\centering
	\begin{subfigure}{0.45\linewidth}
		\centering
		\includegraphics[width=0.95\linewidth]{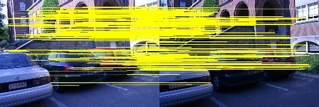}
		\caption{Leuven KT-ORB}
		\label{Fig111}%
	\end{subfigure}
	\centering
	\begin{subfigure}{0.45\linewidth}
		\centering
		\includegraphics[width=0.95\linewidth]{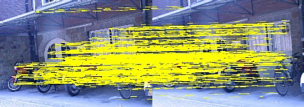}
		\caption{Bikes KT-ORB}
		\label{Fig112}
	\end{subfigure}
	\centering
	\begin{subfigure}{0.45\linewidth}
		\centering
		\includegraphics[width=0.95\linewidth]{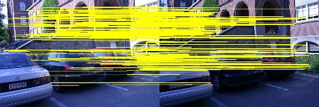}
		\caption{Leuven KTG-ORB}
		\label{Fig113}%
	\end{subfigure}
	\centering
	\begin{subfigure}{0.45\linewidth}
		\centering
		\includegraphics[width=0.95\linewidth]{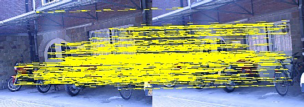}
		\caption{Bikes KTG-ORB}
		\label{Fig114}
	\end{subfigure}
	\centering
	\begin{subfigure}{0.45\linewidth}
		\centering
		\includegraphics[width=0.95\linewidth]{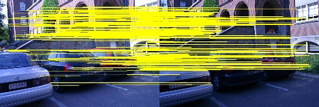}
		\caption{Leuven KTGP-ORB}
		\label{Fig115}%
	\end{subfigure}
	\centering
	\begin{subfigure}{0.45\linewidth}
		\centering
		\includegraphics[width=0.95\linewidth]{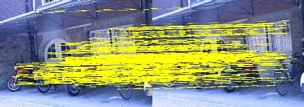}
		\caption{Bikes KTGP-ORB}
		\label{Fig116}
	\end{subfigure}
	\caption{Comparison of experimental results for each model at different stages.}
\end{figure}
  To quantitatively evaluate the models at different stages, feature extraction and matching experiments were conducted on all 5 image pairs of the Leuven and Bikes datasets. As shown in Table 3, for each dataset, matching was performed using different models, and the average NM (Number of Matchings) was calculated. Here, NM represents the remaining number of matches after eliminating erroneous, low-quality, and redundant matches. A lower NM value indicates a higher number of eliminated erroneous, low-quality, and redundant matches.

  With the introduction of KNN, TF, GMS, and PROSAC techniques into the matching models, there is a noticeable decrease in the number of matches for each dataset. As the model transitions from KT-ORB to KTGP-ORB, the average NM values for the Leuven and Bikes datasets decrease from 552.2 to 479.4 and from 603.2 to 525.8, respectively. This implies that through multi-stage refined matching, not only can erroneous or low-quality matches be removed, but redundant matches can also be gradually eliminated.
  \begin{table*}[t]
	\newcommand{\tabincell}[2]{\begin{tabular}{@{}#1@{}}#2\end{tabular}}
	\centering
	\caption{NM of each model at different stages. }
	\setlength{\tabcolsep}{2.0pt}
	\begin{tabular}{ccccccc}
	\toprule
	 \tabincell{c}{Image Pairs} & \multicolumn{2}{c}{Leuven} & \multicolumn{2}{c}{Bikes} \\
	\cmidrule(r){2-3} \cmidrule(lr){4-5} 
	&KT-ORB &  \tabincell{c}{KTG-ORB}  & KTGP-ORB & KT-ORB & KTG-ORB & \tabincell{c}{KTGP-ORB} \\
	\midrule
	1-2	&833	&791	&785	&1019	&934	&928\\
	1-3	&634	&550	&546	&843	&768	&767\\
	1-4	&514	&406	&405	&525	&459	&456\\
	1-5	&421	&352	&352	&372	&338	&330\\
	1-6	&359	&315	&309	&257	&206	&198\\
	Mean	&552.2	&482.8	&479.4	&603.2	&541	&525.8\\

	\bottomrule
	\end{tabular}
	\label{tab3}
	\end{table*} 
  Moreover, matching accuracy is a crucial metric for feature extraction and matching. Therefore, an analysis of the REP (Repeatability), ME (Mean Error), and RMSE (Root Mean Square Error) evaluation metrics for different models was conducted. Feature extraction and matching were performed on 5 pairs of images for both datasets, and the average REP was calculated. Figure 9 illustrates the average REP values for different models.

  It is evident from the figure that the KTG-ORB model, which incorporates GMS, exhibits a significant increase in REP values compared to the KT-ORB model for both datasets. This improvement is primarily attributed to the fact that KNN and TF only consider the similarity of ORB features based on local appearance in the Hamming space, inevitably leading to a certain number of erroneous matches. With the integration of GMS into the model, considering the motion smoothness constraint of local images, a substantial number of mismatches and low-quality matches are eliminated. Furthermore, the integration of PROSAC into the model results in a slight improvement in REP values. The fundamental reason can be attributed to the fact that GMS primarily rapidly determines whether a match is correct or incorrect, leaving some redundant matches. This implies that KTGP-ORB not only utilizes Hamming distance but also employs Euclidean distance to eliminate erroneous, low-quality, and redundant 

  \begin{figure}[t]
    \centering
    \includegraphics[height=2.5in]{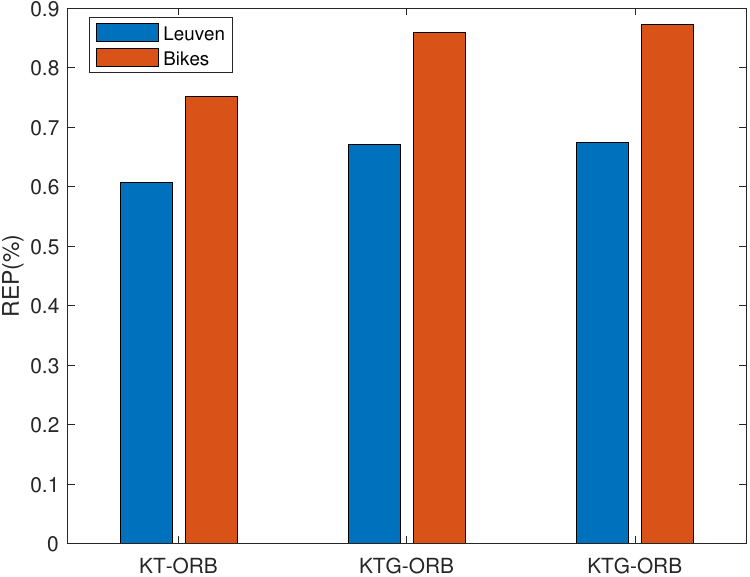}
\caption{Average REP values for different models.}
    \label{Fig12}
\end{figure}

  Further calculations were conducted to determine the average ME (Mean Error) and RMSE (Root Mean Square Error) values for different models on the five pairs of images from the two datasets, as presented in Table 4 and Table 5. The results indicate that the KTGP-ORB model exhibits the highest matching accuracy for both the Leuven and Bikes datasets. When transitioning from the KT-ORB model to the KTGP-ORB model, both ME and RMSE values decrease. For instance, in the Leuven dataset, ME decreases from 23.85192 pixels to 27.73106 pixels, and RMSE decreases from 25.37288 pixels to 25.23356 pixels.

  Therefore, the KTGP-ORB model can obtain an ample number of matches from these two datasets with high matching accuracy. This validates the effectiveness of the KNN-TF-GMS-PROSAC (KTGP) multi-stage refined matching model in excluding erroneous and redundant matches while retaining high-quality matches, particularly in challenging scenarios with factors like varying lighting conditions and blurriness.
  
  \begin{table}[ht]
	\newcommand{\tabincell}[2]{\begin{tabular}{@{}#1@{}}#2\end{tabular}}
	  \centering
	  \caption{Average ME values for different models.}
	  \setlength{\tabcolsep}{4pt} 
	  \begin{tabular}{ccc}
	  \toprule
	  \multicolumn{1}{c}{Algorithm Model} & \multicolumn{2}{c}{ME (pixels)}  \\
	  \cmidrule(lr){2-3} 
	  & Leuven & Bikes  \\
	  \midrule
	  KT-ORB	& 23.85192 & 24.11962 \\
	  KTG-ORB	& 23.83598 & 24.08338 \\
	  KTGP-ORB & 23.73106 & 23.99844 \\	
	  \bottomrule
	  \end{tabular}
	  \label{tab4}
  \end{table}

	\begin{table}[ht]
		\newcommand{\tabincell}[2]{\begin{tabular}{@{}#1@{}}#2\end{tabular}}
		\centering
		\caption{Average RMSE values for different models. }
		\setlength{\tabcolsep}{4pt}
		\begin{tabular}{ccc}
		\toprule
		 {Algorithm  Model} & \multicolumn{2}{c}{RMSE (pixels)}  \\
		\cmidrule(lr){2-3} 
		& Leuven &  \tabincell{c}{Bikes}  \\
		\midrule
		KT-ORB	&25.37288	&25.68636\\
		KTG-ORB	&25.35014	&25.64750\\
		KTGP-ORB	&25.23356	&25.56006\\

		\bottomrule
		\end{tabular}
		\label{tab5}
		\end{table}

  The above experimental results indicate that KTGP-ORB is the optimal model across different stages. To further showcase the advantages of the KTGP-ORB model, a comparison was made between KTGP-ORB and the classical ORB algorithm on the five pairs of images from the two datasets. The average values of REP, ME, and RMSE were calculated, as presented in Table 6.

  The KTGP-ORB model demonstrates superior stability in challenging scenarios with varying lighting conditions and blurriness when compared to the ORB algorithm. The average REP values for KTGP-ORB on the Leuven and Bikes datasets are 0.67 and 0.87, respectively, while the ORB algorithm only achieves 0.37 and 0.45. This suggests that the KTGP-ORB model is more stable in the presence of lighting variations and blurriness.

  Additionally, the average ME and RMSE values for KTGP-ORB on the Leuven and Bikes datasets are 23.73106, 25.23356, and 23.99844, 25.56006, respectively. In comparison, the ORB algorithm yields average ME and RMSE values of 33.90576, 37.02068, and 34.1985, 36.83376 on the Leuven and Bikes datasets, respectively. Consequently, the KTGP-ORB model achieves an average reduction of 29.92
  \begin{table}[h]
	\newcommand{\tabincell}[2]{\begin{tabular}{@{}#1@{}}#2\end{tabular}}
	\centering
	\caption{Average values of KTGP-ORB and ORB under different evaluation metrics. }
	\setlength{\tabcolsep}{2.0pt}
	\begin{tabular}{ccccc}
	\toprule
	 \tabincell{c}{Metric} & \multicolumn{2}{c}{Leuven} & \multicolumn{2}{c}{Bikes} \\
	\cmidrule(r){2-3} \cmidrule(lr){4-5} 
	& ORB &  \tabincell{c}{KTGP-ORB}  & ORB &  \tabincell{c}{KTGP-ORB} \\
	\midrule
	REP (\%)	&0.37	&0.67	&0.45	&0.87\\
	ME (pixels)	&33.90576	&23.73106	&34.1985	&23.99844\\
	RMSE (pixels)	&37.02068	&25.23356	&36.83376	&25.56006\\	
	\bottomrule
	\end{tabular}
	\label{tab6}
	\end{table} 
  \section{Conclusion}
  This paper proposes a novel KTGP-ORB model based on the KNN-TF-GMS-PROSAC (KTGP) multi-stage fine matching technique. The model leverages the similarity of ORB feature pairs in the Hamming space, considering the local appearance similarity of feature descriptors. To address issues such as ambiguity, false matches, and the reduction in accuracy caused by low initial matching precision, the model introduces constraints based on local image motion smoothness to enhance the accuracy of PROSAC sampling in the initial matching phase. Furthermore, the global grayscale information of feature pairs in the Euclidean space is optimized using the PROSAC algorithm. Experimental results in challenging scenarios with lighting variations and blurriness demonstrate the effectiveness of the KNN-TF-GMS-PROSAC (KTGP) multi-stage fine matching technique. It successfully excludes erroneous matches and redundant matches while preserving high-quality matches. Comparative experiments with the classical ORB algorithm highlight the superiority of the KTGP-ORB model, showing an average reduction of 29.92\% in errors in scenarios involving lighting variations and blurriness.





\end{document}